\documentclass[10pt,twocolumn,letterpaper]{article}

\usepackage[pagenumbers]{cvpr}

\usepackage{subcaption}
\usepackage{multirow}
\usepackage{arydshln}
\usepackage{amsfonts}
\usepackage{booktabs}

\definecolor{cvprblue}{rgb}{0.21,0.49,0.74}
\usepackage[pagebackref,breaklinks,colorlinks,allcolors=cvprblue]{hyperref}

\title{Characterizing Motion Encoding in Video Diffusion Timesteps}

\author{
Vatsal Baherwani\thanks{Equal Contribution.}\quad
Yixuan Ren\footnotemark[1]\quad
Abhinav Shrivastava\\
University of Maryland
}

\begin{document}

\maketitle

\begin{abstract}

Text-to-video diffusion models synthesize temporal motion and spatial appearance through iterative denoising, yet how motion is encoded across timesteps remains poorly understood. Practitioners often exploit the empirical heuristic that early timesteps mainly shape motion and layout while later ones refine appearance, but this behavior has not been systematically characterized. In this work, we proxy motion encoding in video diffusion timesteps by the trade-off between appearance editing and motion preservation induced when injecting new conditions over specified timestep ranges, and characterize this proxy through a large-scale quantitative study. This protocol allows us to factor motion from appearance by quantitatively mapping how they compete along the denoising trajectory. Across diverse architectures, we consistently identify an early, motion-dominant regime and a later, appearance-dominant regime, yielding an operational motion-appearance boundary in timestep space. Building on this characterization, we simplify current one-shot motion customization paradigm by restricting training and inference to the motion-dominant regime, achieving strong motion transfer without auxiliary debiasing modules or specialized objectives. Our analysis turns a widely used heuristic into a spatiotemporal disentanglement principle, and our timestep-constrained recipe can serve as ready integration into existing motion transfer and editing methods.
\end{abstract}    
\section{Introduction}

\begin{figure}[!t]
    \centering
    \includegraphics[width=\linewidth]{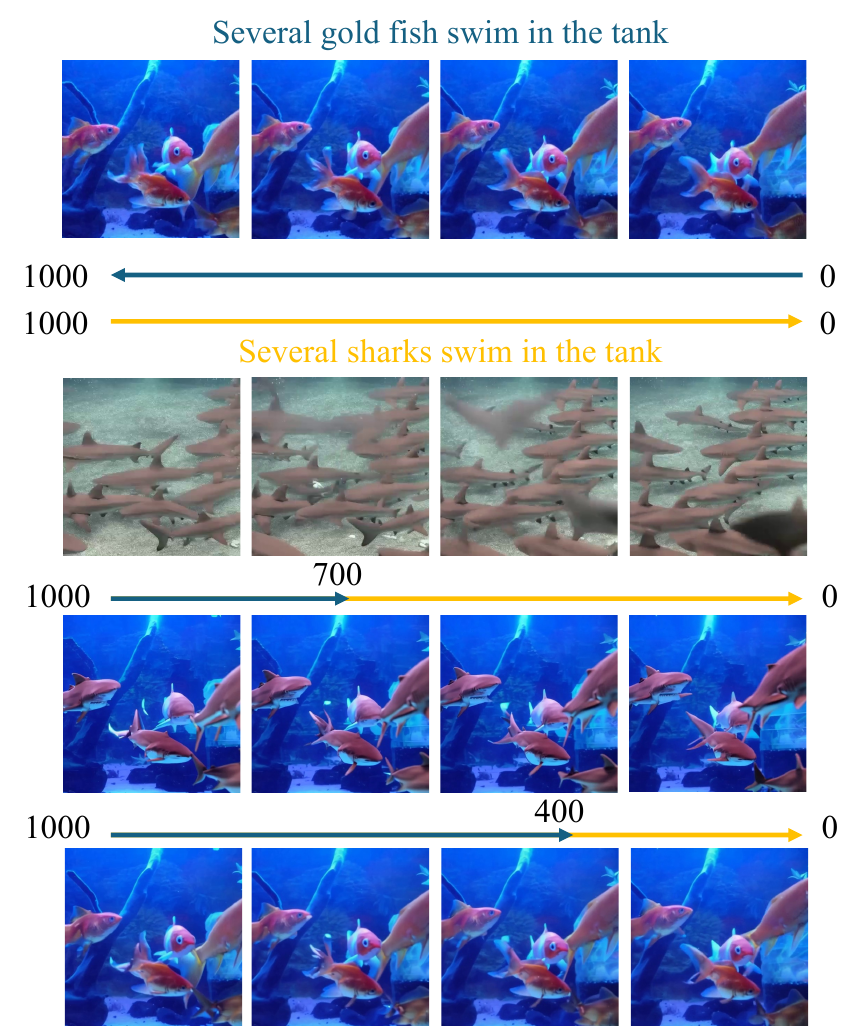}
    \caption{Spatiotemporal disentanglement in video diffusion models.
    Our finding reveals that motion is primarily encoded in the early denoising timesteps.
    Given a reference video (top) and its ground truth caption (blue), we perform DDIM inversion and then denoise with a new prompt that modifies only the subject (yellow).
    The resampled videos show different subject editing and motion preservation results by applying the original or new prompts at different timesteps.
    }
    \label{fig:teaser}
\end{figure}

Diffusion models \cite{ho2020denoising} have achieved remarkable performance in image and video synthesis, and large-scale pre-trained foundations now support many controllable generation tasks such as editing \cite{zhang2023adding, tumanyan2023plug} and customization \cite{ruiz2023dreambooth, gal2022imageworthword}. For videos, a key challenge is that temporal motion and spatial appearance are entangled in the generative process, while many applications require modifying only one factor and preserving the other. Understanding how motion is encoded along the denoising trajectory is therefore central to controllable video generation.

Recent work has analyzed what different timesteps and layers represent in image diffusion models, revealing coarse-to-fine synthesis and frequency-specific behaviors \cite{hertz2022prompt, luo2023diffusion, yue2024exploring, qian2024boosting, lee2025beta, liang2025aesthetic, wu2025freediff}. In the video domain, several methods \cite{xiao2024video, li2024personalvideo, wu2025customcrafter} empirically exploit that early denoising steps tend to determine motion and layout, while later steps refine appearance. This heuristic has been used to design subject editing and motion transfer pipelines, but it remains informal: there is no systematic quantification of how motion and appearance trade off across timesteps, how consistent this pattern is across architectures and datasets, or where an operational boundary between motion-dominant and appearance-dominant regimes lies.

In this work, we provide an in-depth characterization of motion encoding across timesteps in text-to-video diffusion models. Given a reference video and its text prompt, we obtain a denoising trajectory and resample the video while replacing only the appearance-related part of the prompt over specified ranges of timesteps, keeping the original prompt at other steps. Although this procedure is not intended as a high-quality editing method, it produces tampered RGB videos that reveal how strongly appearance can be edited and how well the original motion is preserved when re-conditioning at different timesteps. By measuring appearance alignment and motion preservation, we use the resulting trade-off as a large-scale quantitative proxy for how motion is encoded in video diffusion timesteps.

Sweeping over timestep ranges reveals a consistent spatiotemporal structure: an early motion-dominant regime where re-conditioning strongly affects temporal dynamics, and a later appearance-dominant regime where re-conditioning mainly changes spatial details while largely keeping the motion. This pattern induces an operational motion-appearance boundary in timestep space, defined as the range where appearance editing becomes effective yet motion preservation remains high. We validate this behavior across three text-to-video architectures: ModelScope \cite{modelscope}, a U-Net with dedicated spatial and temporal attention blocks; Latte \cite{ma2024latte}, a Transformer with decoupled spatial and temporal attentions; and CogVideoX \cite{yang2024cogvideox}, a Transformer with unified spatiotemporal attention. Despite their architectural varieties, all models display similar motion-dominant and appearance-dominant regimes. 

Based on this characterization, we derive a simplified design principle for motion-centric adaptation: restrict modeling to motion-dominant timesteps. We instantiate this principle in a one-shot video motion customization framework, where a single reference video provides the target motion that should be transferred to new subjects and scenes with temporal diversity. Prior works typically introduce auxiliary debiasing modules or specialized losses to suppress unwanted spatial signals \cite{zhao2023motiondirector, ren2024customizeavideo}. In contrast, we fine-tune temporal attention in pre-trained text-to-video diffusion models using the vanilla diffusion loss, while constraining both training and inference to early timesteps. This timestep constraint effectively prevents appearance leakage despite using full reconstruction losses, and it enables efficient partial-attention tuning and even direct full-rank fine-tuning without triggering spatial overfitting.

In summary, our main contributions are:
\begin{itemize}
    \item We introduce a prompt-tampering probe and quantitatively analyze the trade-off between appearance editing and motion preservation across timesteps in pre-trained text-to-video diffusion models.
    \item We identify consistent motion-dominant and appearance-dominant regimes across diverse architectures and characterize an operational motion-appearance boundary in timestep space.
    \item Guided by this boundary, we propose a timestep-constrained one-shot motion customization framework that requires no auxiliary debiasing modules and naturally supports partial-attention tuning and direct tuning.
\end{itemize}
\section{Related Works}

\subsection{Diffusion Attribute Disentanglement}

Attribute disentanglement in diffusion models is increasingly studied as a way to interpret internal representations and gain finer control. For image generation, several works analyze how information is organized across timesteps and layers. Aggregating multi-timestep and multi-scale features reveals complementary geometric and semantic cues for correspondence \cite{Luo2023DiffusionHyperfeatures}, and spectral analyses show that low-frequency content dominates early steps while high-frequency refinements appear later, motivating non-uniform timestep sampling and frequency-aware manipulation \cite{Lee2024BetaSampling,wu2025freediff}. Other methods make timesteps explicit supervision axes through timestep-aware representations and step-aware preference alignment \cite{Yue2024DiTi,Chen2024SPO,Sun2024DDE}, while per-step editing demonstrates that intervening at selected timesteps can separate layout from style \cite{Hertz2022PromptToPrompt}. Recent interpretability work further shows that semantic concepts are structured across layers and timesteps \cite{kim2025revelio}. These studies, however, focus on spatial attributes in image diffusion and do not characterize how motion is encoded along timesteps in video models.

For video diffusion models, timestep-wise disentanglement is less developed and mostly used in a heuristic way.
\cite{li2024personalvideo,wu2025customcrafter} inject new appearance into reference videos often bypass early steps to reduce motion interference, implicitly assuming that motion is encoded early and appearance later, but without quantifying where motion and appearance respectively dominate.
\cite{bahmani2025ac3d} studies how camera trajectories are encoded over timesteps and separates camera from scene content, whereas our goal is to understand general object and scene motions and their interaction with appearance.
\cite{zhang2025flexiact} learns frequency-aware embeddings across all timesteps for image-to-video (I2V) models, whose appearance has been mostly debiased by the image image.
\cite{xiao2024video,ling2024motionclone} extract motion-aware features from pre-trained T2V models and guide motion by feature alignment without tuning, while our motion module models the reference motion signal and is able to adapt it to any novel scenarios with temporal diversity.
Moreover, we systematically characterize an architecture-agnostic, quantitative generic motion-appearance boundary that describes how motion is encoded in text-to-video diffusion timesteps.

\subsection{Video Motion Customization}

Video motion customization aims to learn motion from reference videos and transfer it to new subjects and scenarios. Some methods achieve deterministic editing or motion transfer by supplying strong external guidance such as edge or depth maps \cite{chen2023control,zhang2305controlvideo,zhao2023controlvideo}, optical flow \cite{yang2023rerender,liang2024flowvid}, or latent feature alignment \cite{geyer2023tokenflow,ling2024motionclone}. These approaches operate at inference time without fine-tuning the backbone and primarily focus on faithfully following the given control signals.

Another line of work fine-tunes pre-trained text-to-video diffusion models to adopt the desired motion with temporal diversity. In the one-shot regime, appearance and motion are tightly entangled and models tend to overfit appearance. Spatial debiasing modules and tailored objectives are introduced to encourage temporal adapters to focus on motion and suppress appearance leakage \cite{zhao2023motiondirector,ren2024customizeavideo}, while temporal feature losses are designed to distill motion without copying content \cite{wu2025motionmatcher}. These methods rely on auxiliary modules or specialized losses to approximate motion-appearance decoupling. By contrast, our approach starts from a characterization of motion encoding across timesteps: we identify a motion-dominant regime in the denoising schedule and constrain both training and inference to these timesteps, using the vanilla diffusion loss with standard temporal attention adapters. This timestep-based design prevents appearance leakage while enabling flexible motion customization with partial-attention tuning or direct full-rank tuning.

\section{Spatiotemporally Disentangled Diffusion}

\subsection{Preliminary}

\paragraph{Diffusion Models}
Diffusion models \cite{ho2020denoising} generate synthetic instances by sampling $\mathbf{x}_T\sim\mathcal{N}(\mathbf{0}, \mathbb{I})$ and iteratively applying a denoising process to obtain $\mathbf{x}_0$ via
\begin{equation}
\label{eq:sampling}
    \mathbf{x}_{t-1} = 
    \frac{1}{\sqrt{\alpha_t}}
    \left(
    \mathbf{x}_t - 
    \frac{1 - \alpha_t}
    {\sqrt{1 - \bar\alpha_t}}
    \epsilon_\theta(\mathbf{x}_t, t, c)
    \right)
    + \sigma_t\mathbf{z} ,
\end{equation}
where $t=T,...,1$.
$\epsilon_\theta$ is a parameterized denoising neural network with a condition $c$, $\mathbf{z}\sim\mathcal{N}(\mathbf{0}, \mathbb{I})$ is random noise, $\sigma_t$ is the variance, and $\alpha_t, \bar\alpha_t$ are hyperparameters defining the noise schedule.

\paragraph{Text-to-Video Diffusion Models}
In text-to-image diffusion models, $c$ is a text prompt depicting the expected output video, and a typical $\epsilon_\theta$ comprises self-attentions and cross-attentions to process the visual information with the condition incorporated.
To synthesize sequential data consisting of multiple images, $\epsilon_\theta$ additionally involves cross-frame attentions to regularize the temporal consistency.

\paragraph{DDIM Inversion}
In implicit diffusion models (DDIMs, \citeauthor{song2022denoisingdiffusionimplicitmodels}), the denoising process in Eq. \ref{eq:sampling} can be made deterministic by setting $\sigma_t:=0$.
Then the denoising process can be inverted by expressing $x_t$ in terms of $x_{t-1}$ \cite{mokady2022nulltextinversion}, and ultimately producing from an existing $\mathbf{x}_0$ its approximate sampling trajectory $\mathbf{x}_{\{T,...,1\}}$, which reconstructs itself following the denoising process.

\begin{figure}[t]
	\begin{subfigure}[b]{0.31\linewidth}
    \centering
    \includegraphics[width=\textwidth]{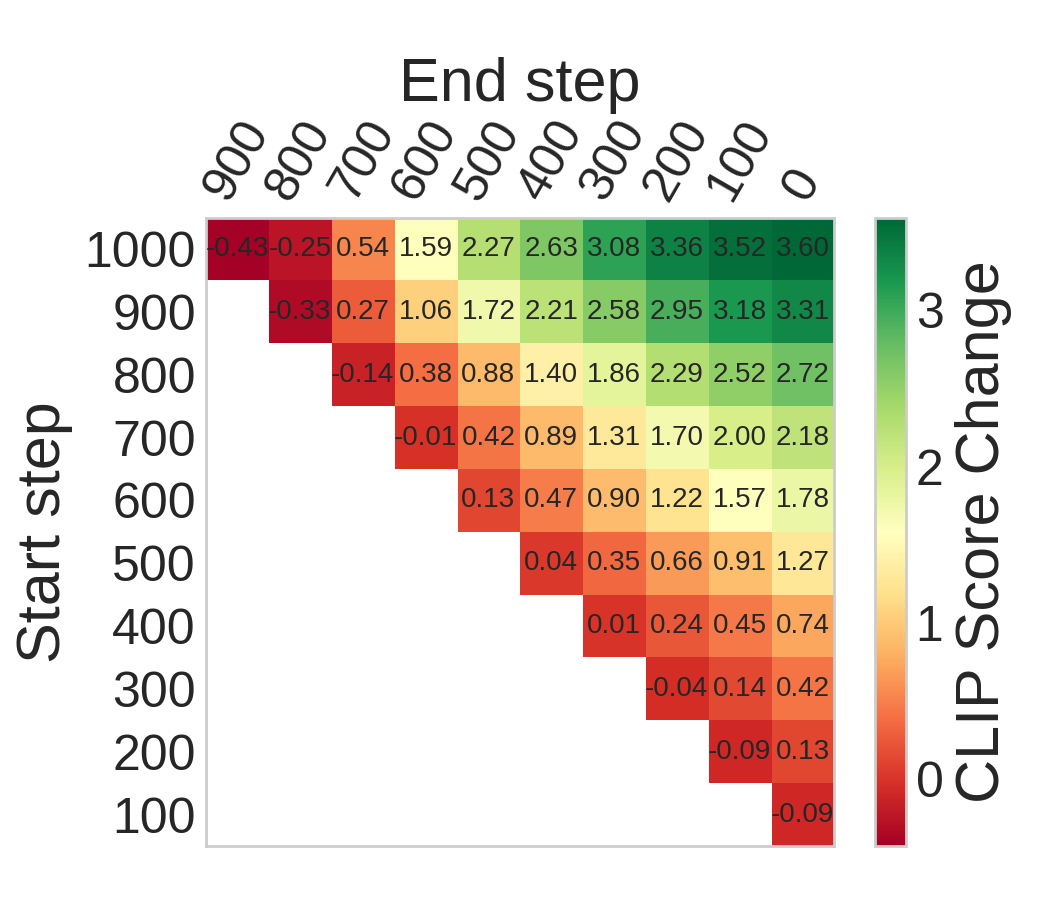}
    \caption{Subject Edit.}
	\end{subfigure} \hfill
	\begin{subfigure}[b]{0.31\linewidth}
    \centering
    \includegraphics[width=\textwidth]{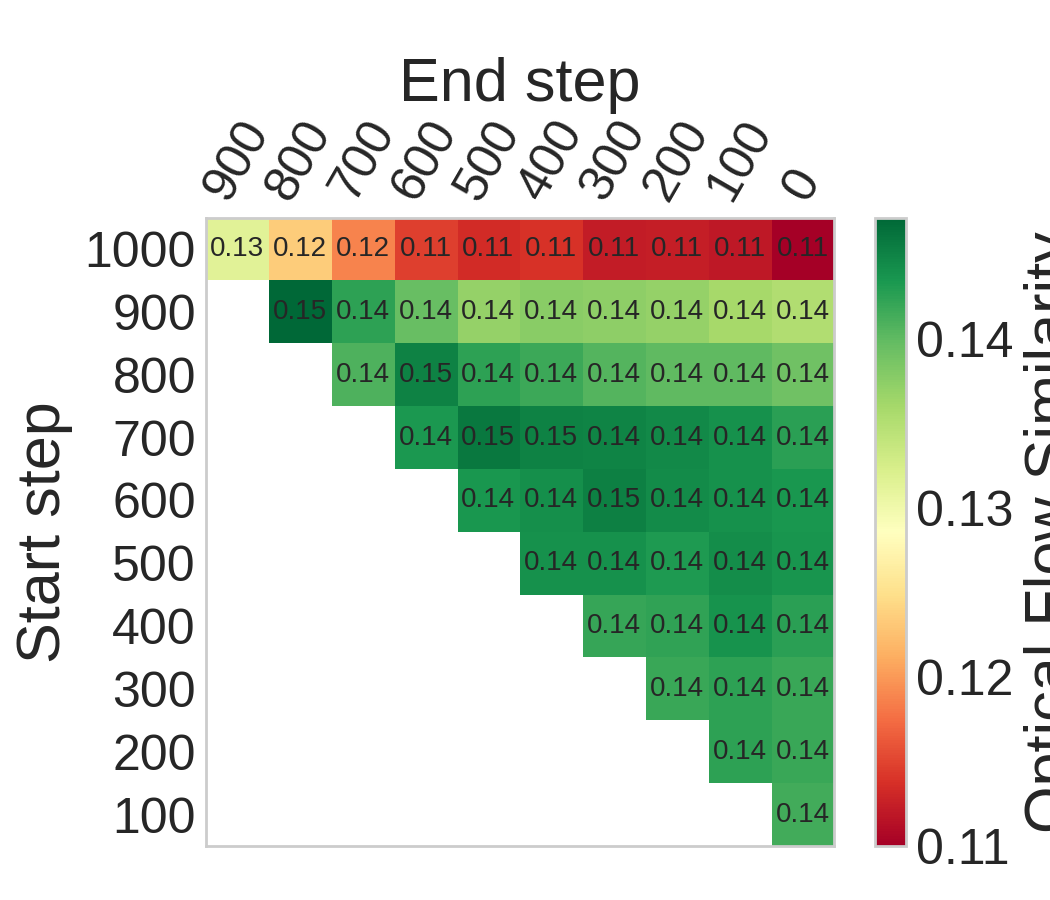}
    \caption{Motion Preserve.}
	\end{subfigure} \hfill
    \begin{subfigure}[b]{0.31\linewidth}
    \centering
    \includegraphics[width=\textwidth]{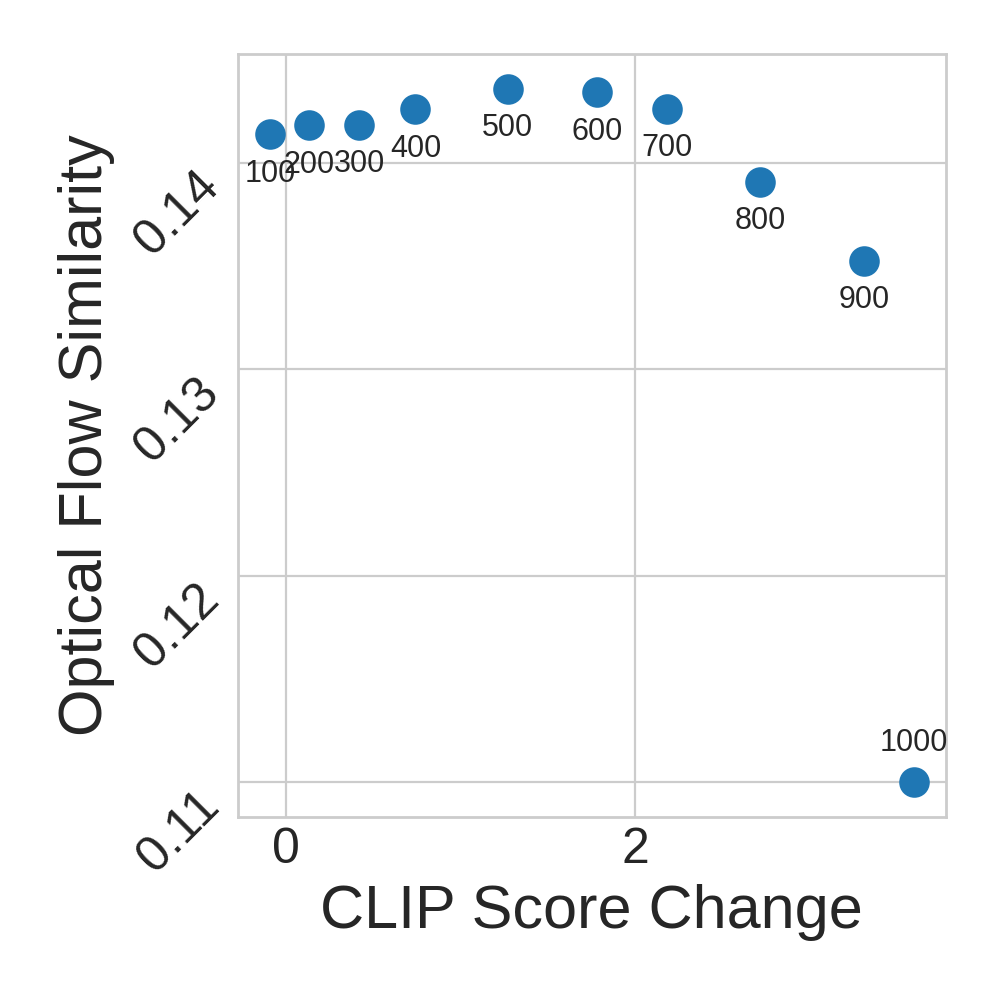}
    \caption{ModelScope}
	\end{subfigure} \hfill
	\begin{subfigure}[b]{0.31\linewidth}
    \centering
    \includegraphics[width=\textwidth]{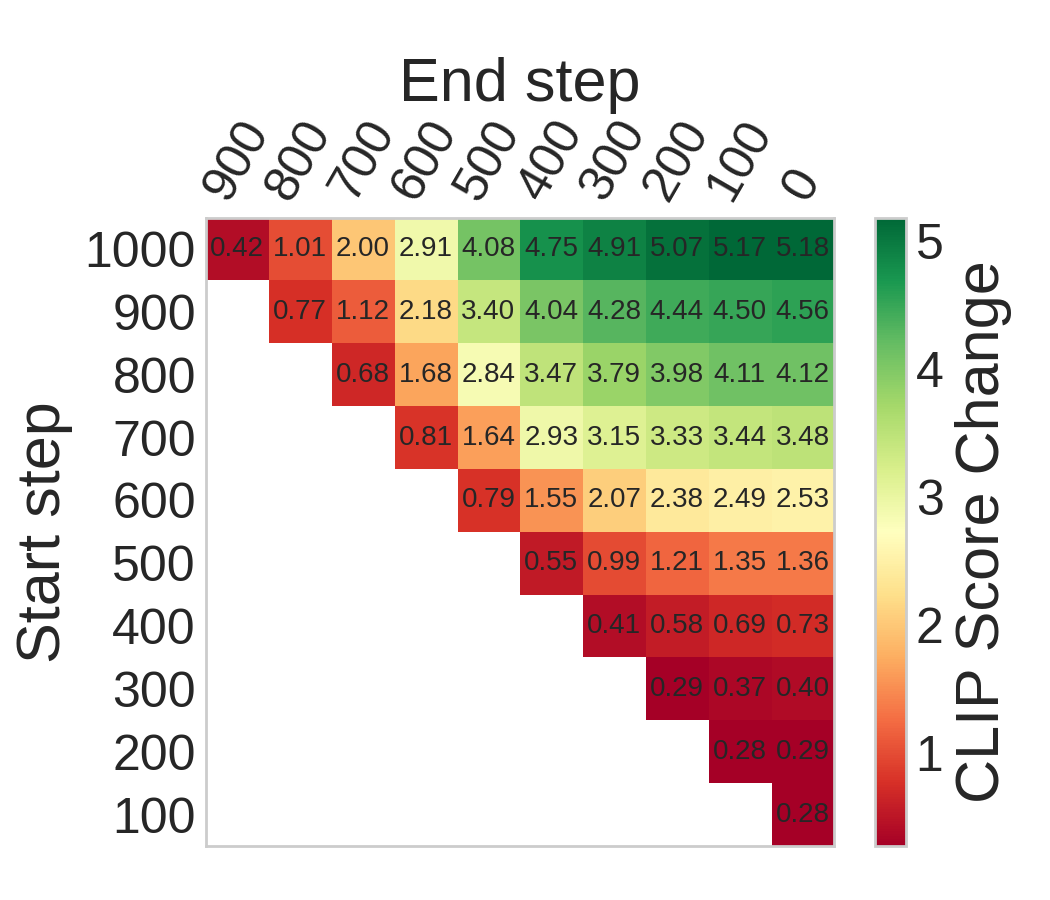}
    \caption{Subject Edit.}
	\end{subfigure} \hfill
	\begin{subfigure}[b]{0.31\linewidth}
    \centering
    \includegraphics[width=\textwidth]{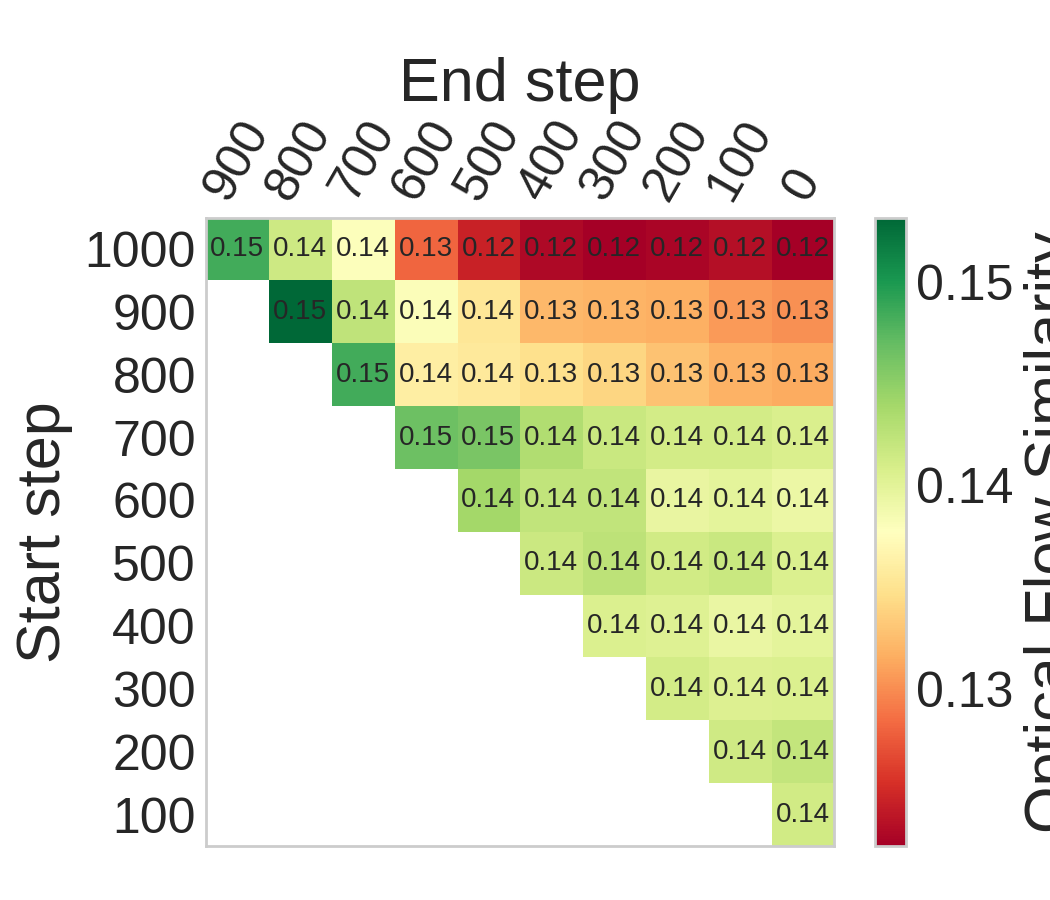}
    \caption{Motion Preserve.}
	\end{subfigure} \hfill
	\begin{subfigure}[b]{0.31\linewidth}
    \centering
    \includegraphics[width=\textwidth]{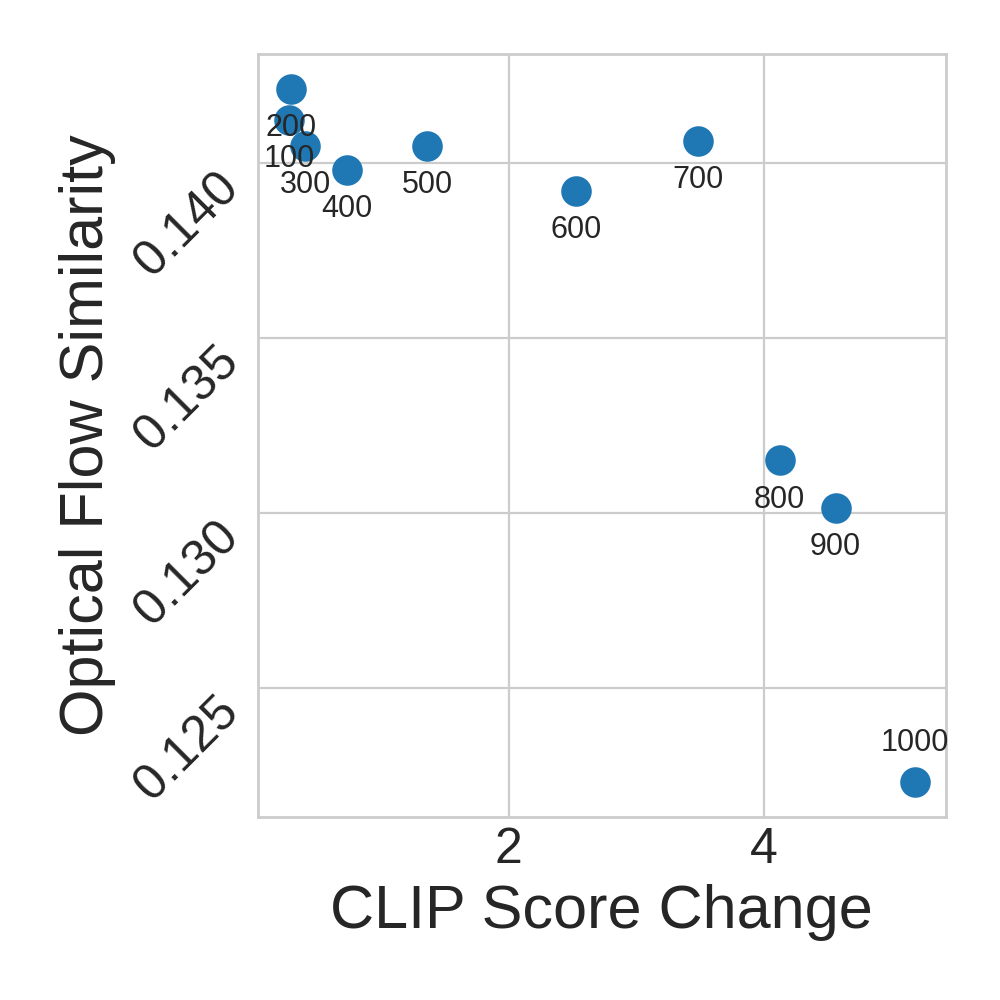}
    \caption{Latte}
	\end{subfigure} \hfill
	\begin{subfigure}[b]{0.31\linewidth}
    \centering
    \includegraphics[width=\textwidth]{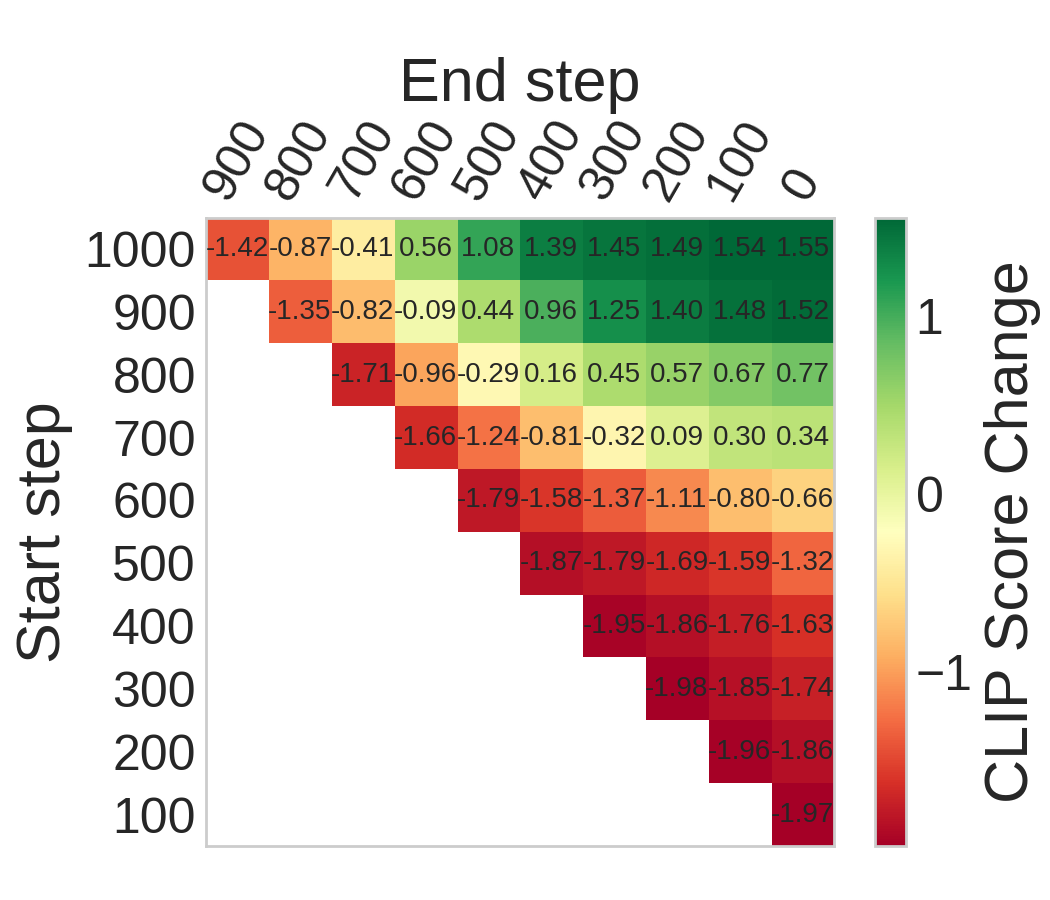}
    \caption{Subject Edit.}
	\end{subfigure} \hfill
	\begin{subfigure}[b]{0.31\linewidth}
    \centering
    \includegraphics[width=\textwidth]{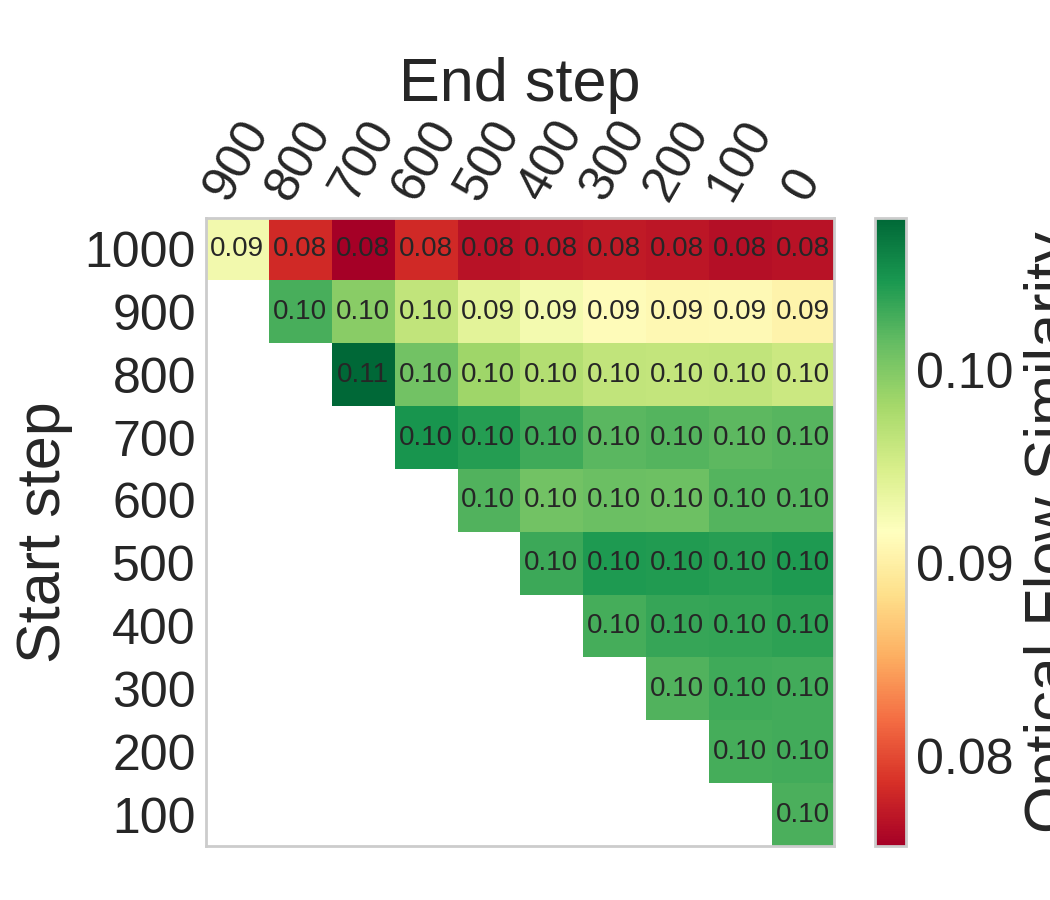}
    \caption{Motion Preserve.}
	\end{subfigure} \hfill
	\begin{subfigure}[b]{0.31\linewidth}
    \centering
    \includegraphics[width=\textwidth]{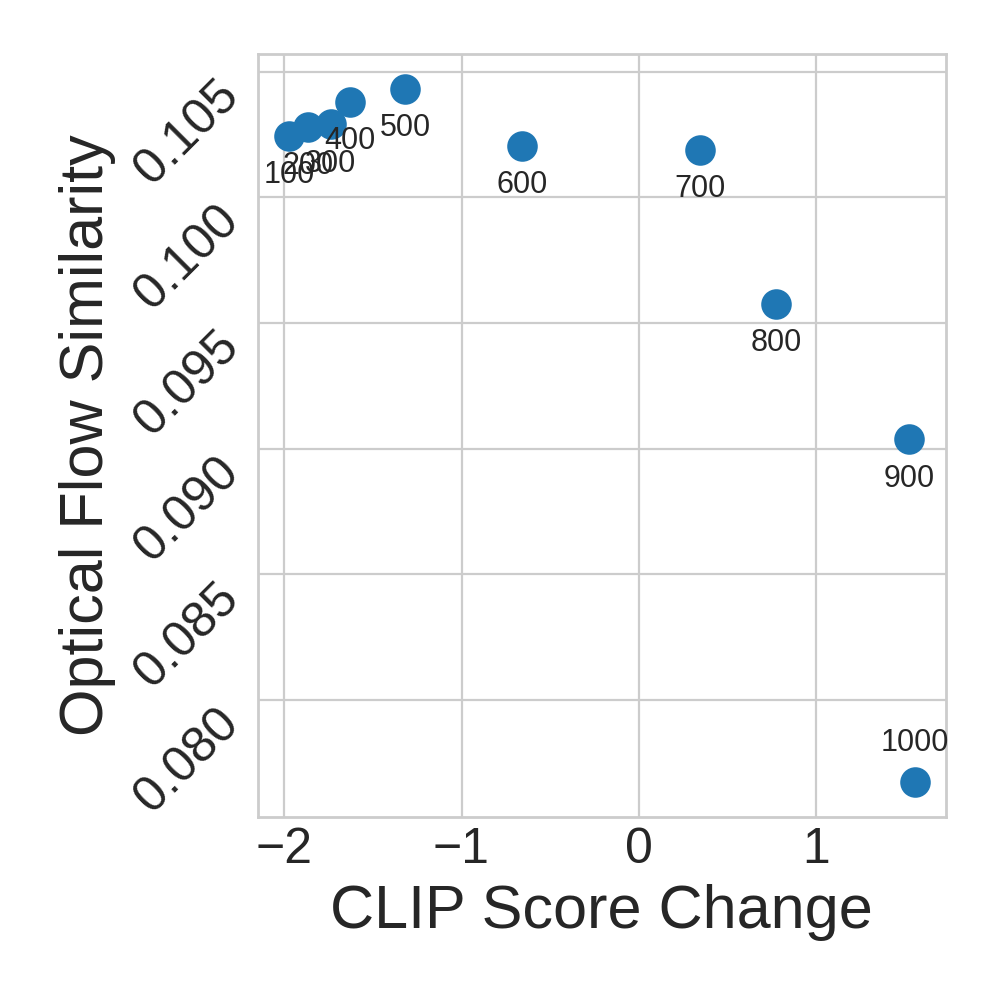}
    \caption{CogVideoX}
	\end{subfigure}
	\caption{Subject editing and motion preservation quality of ModelScope, Latte and CogVideoX.
    Applying the new subject editing prompt in longer timesteps always leads to stronger new subject representation in the generated video.
    However, starting resampling with the new prompt at early timesteps significantly harms the motion preservation although it doesn't modify the motion description.
    The trade-off curves show the optimal timesteps to decompose spatial and temporal signals.
    This spatiotemporal property holds consistently across different model architectures.
    }
    \label{fig:metrics}
\end{figure}

\subsection{Analysis Design}
We aim to observe how the spatial and temporal attributes of a video are processed at various timesteps in the diffusion and denoising processes.
However, this is not trivial as categorizing appearance and motion can be ambiguous in general.
And understanding from noisy videos at intermediate diffusion timesteps further lifts its difficulty.
Therefore, we design to leverage the inversion approach and tamper the resampling trajectory for feasible calculation and reference.

Specifically, given a video $x_{0}$ and its ground truth caption $c$, we start from DDIM inversion to acquire its noise latent $\hat{x}_{T}$, such that the denoising network $\theta$ can faithfully recover it via the original trajectory $x_{0}=\prod_{t=T}^{1} \theta(\hat{x}_{t}|t, c)$.
Next, we tamper of $c$ to $c'$ by changing its subject, and perform denoising process with the edited condition $x'_{0}=\prod_{t=T}^{1} \theta(\hat{x}_{t}|t, c')$.
While $x'_0$ is ideally expected to represent the new subject with the original motion preserved as indicated by $c'$, this process will in fact intervene the generated motion as well, as show in Fig. \ref{fig:teaser} row 2.

Based on this, we propose to examine how the denoising timesteps interact with the new text prompt to synthesize new appearance and original motion.
To this end, we perform the resampling process with $c'$ in a certain timestep range, and the original $c$ is used outside, as shown in Fig. \ref{fig:teaser} rows 3 and 4.
Formally, we denoise via $x''_{0}=\prod_{t=T}^{1} \theta(\hat{x}_{t}|t, c''_{t})$, where $c''_{t}=c'$ when $t\in[\tau_{\mathrm{start}}, \tau_{\mathrm{end}}]$ and otherwise $c''_{t}=c$.
Then we measure the appearance editing by the CLIP score  \cite{hessel2021clipscore} between $x''_{0}$ and $c'$, and measure the motion preservation by the optical flow similarity between $x''_{0}$ and $x_{0}$.

In this way, we leverage the text captions as comprehensive spatiotemporal labels that are clear and easy to manipulate, and obviate direct calculations on noisy videos or compare across different noise levels via diffusion inversion and resampling in clean latent distribution.
Note that although this naive resampling is not able to perfectly edit the original video reasonably and realistically, it can serve as an analytic approach to exhibit the difference in spatial and temporal impact across timesteps in our evaluation.

\subsection{Experiment Setup}

We consider full combination of all valid $(\tau_{start}, \tau_{\mathrm{end}})$ pairs with an interval of $100$ over the whole $1000$ timesteps.
A visual example of this approach is shown in Fig. \ref{fig:teaser}. Here we use start timestep $\tau_{start}=700$ and end timestep $\tau_{\mathrm{end}}=0$.
As a result, our newly generated video preserves the information from $t\in[700, 1000]$ in the original video.

To fully reflect the editing improvement, we meaure the CLIP score change where the base score between $x_{0}$ and $c'$ is subtracted, as $x_{0}$ already have some resemblance to $c'$ except the tampered subjects.
We use the Lucas-Kanade method for optical flow estimation, and calculate the average cosine similarity between the normalized vectors of all frames.
Both metrics are higher when the new video $x'_0$ better represents the new subject in $c'$ or better preserves the original motion in $x_0$.

We conduct this experiments on three representative text-to-video models with divergent denoising network architectures: ModelScope \cite{modelscope} with U-Net and dedicated spatial and temporal attentions, Latte \cite{ma2024latte} with transformer and dedicated spatial and temporal attentions, and CogVideoX \cite{yang2024cogvideox}, with transformer and unified spatiotemporal attentions.
We test on all $76$ videos from the Text-Guided Video Editing (TGVE) competition dataset \cite{loveu}, which also provides subject editing captions.

\subsection{Results}

In Fig. \ref{fig:metrics} we show the trade-off between CLIP score change and optical flow similarity across all $(\tau_{start}, \tau_{\mathrm{end}})$ options.
The CLIP score change consistently improves whenever the editing interval $\tau_{start}-\tau_{\mathrm{end}}$ is longer, as this allows for more sampling steps with the new prompt $c'$.
Notably, for any given $\tau_{start}$, the optimal $\tau_{\mathrm{end}}$ is always $0$. 
However, $\tau_{\mathrm{end}}$ does not matter as much for motion preservation.
On the contrary, the optical flow similarity increases as we delay the sampling process to start from later timesteps.
In other words, sampling with the new condition $c'$ at earlier timesteps, harms much its optical flow similarity to the original video despite $c'$'s only modification on the subject.
Based on the observed effect of the subject editing prompt of motion deviation from the original video, we claim that motion signals are dominantly encoded in early denoising timesteps in video diffusion models.

We draw the heatmaps of the appearance editing and motion preservation quality in Fig. \ref{fig:metrics}.
We can deduce from it the dominant ranges of motion and appearance along the denoising timesteps for each pre-trained model.
$\tau_{\mathrm{end}}$ is not significant for motion preservation while being optimal for appearance editing at $0$, at which we therefore fix the end timestep.
Given $\tau_{\mathrm{end}}:=0$, varying the start timestep $\tau_{start}$ presents a trade-off between representing the new subject and retaining the original motion.
That is, $\tau_{start}$ reflects the threshold of denoising timesteps where temporal and spatial signals are encoded.
This tradeoff is also depicted in Fig. \ref{fig:metrics} for each base model.
A smaller $\tau_{start}$ leads to minimal shift in optical flow similarity, while CLIP score improves significantly.
A bigger $\tau_{start}$ results in drastic loss in the motion information from the original video.
The threshold timestep for spatiotemporal disentanglement thus lies somewhere along the Pareto frontier.
In following sections we denote $\tau=\tau_{start}$ as this threshold.
While its exact value varies across specific models, it is consistently around $[700, 900]$.
Next, we demonstrate our spatiotemporal disentanglement property in the downstream application of one-shot video motion customization task.

\begin{figure}[t]
    \centering
    \includegraphics[width=\linewidth]{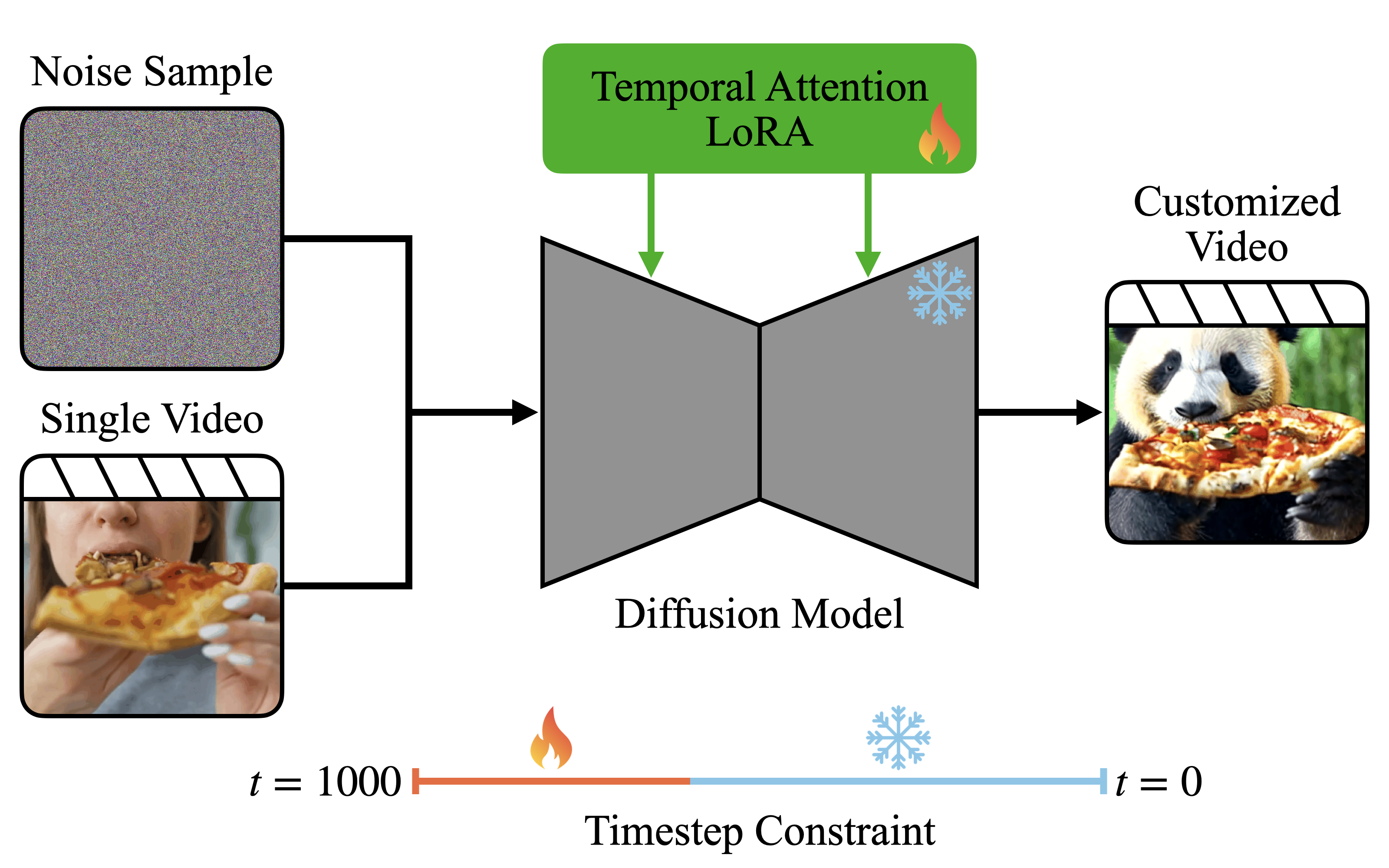}
    \caption{One-shot video motion customization via denoising timestep constraint.
    Leveraging our spatiotemporal disentanglement property, we train LoRAs at only early denoising timesteps to model the reference motion without appearance leakage.
    This single-stage fine-tuning approach achieves surpassing performance without any additional debiasing modules, stages or losses.
    This even works for base models with unified spatiotemporal attentions, where we add LoRA on the full spatiotemporal sequence and it is still prevented from overfitting on the reference appearance.
    }
    \label{fig:method}
\end{figure}

\begin{figure*}[t]
	\centering
	\includegraphics[width=1.05\linewidth]{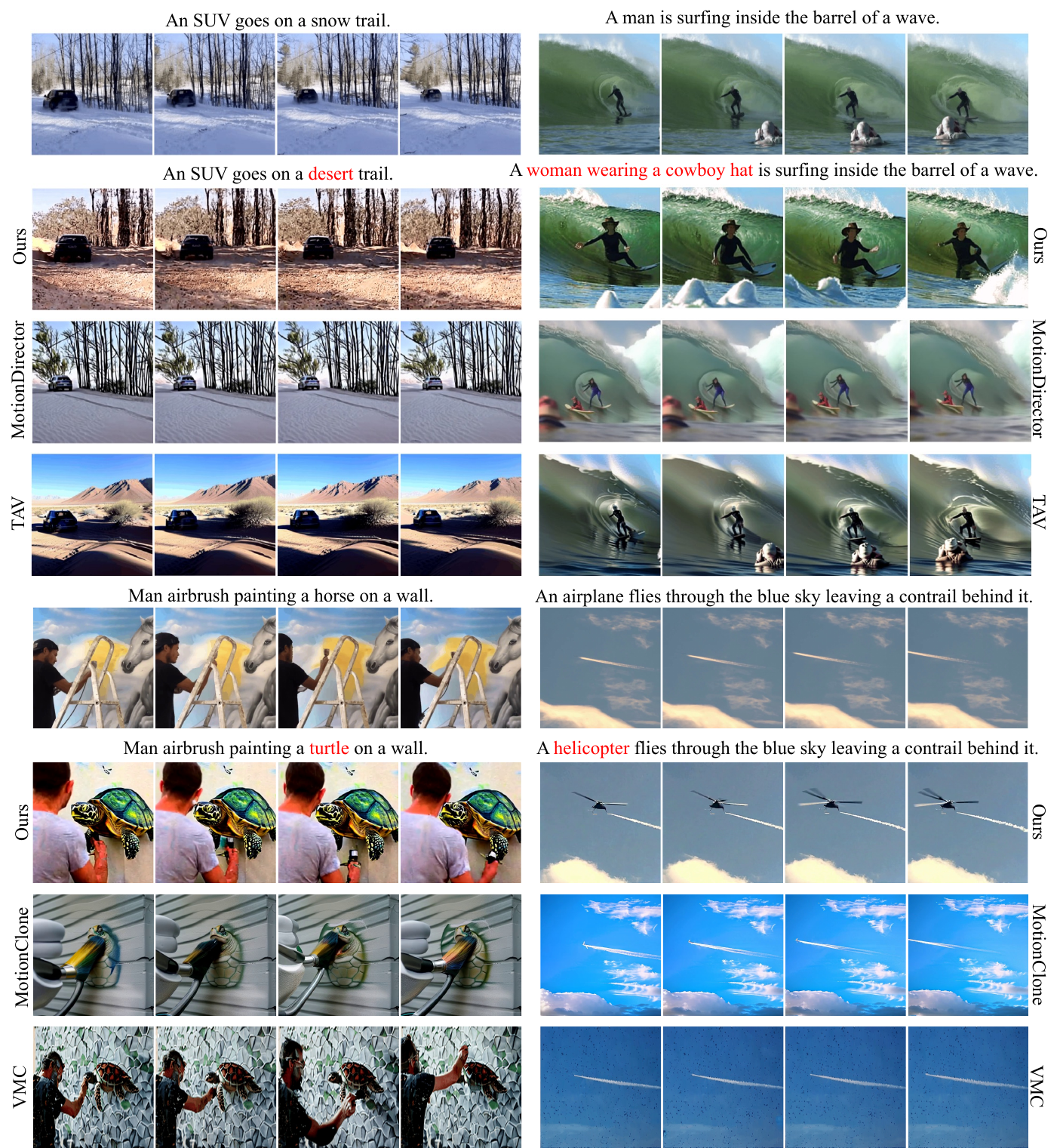}
	\caption{Qualitative comparison of our motion disentanglement method to previous SOTAs.
    Our method faithfully replicates the motion of the reference video while also editing the subject and background with superior quality to other approaches.
    Without any additional spatial debiasing modules or stages, our method is stable and robust with minimal semantic discrepancy (e.g. the snow ground and hat-like reef by MotionDirector, and the extra wall texture and missing object by MotionClone).
    }
    \label{fig:comparison}
\end{figure*}

\section{One-Shot Video Motion Customization}

\subsection{Task Settings}
Video motion customization is the task to customize a pre-trained text-to-video diffusion model with specific motions from given reference videos.
Given the ambiguity of text prompt control of temporal movements, motion customization is the optimal way to replicate the exemplar motions with new subjects and scenes.
Previous methods of video editing and motion transfer aim at generating deterministic movements with precise frame-wise alignment, losing temporal diversities such as motion velocity, intensity, subject count and position, and camera perspective etc.
In contrast, motion customization demands tuning-based modeling of the desired motions and leads to reproducing them with temporal varieties, and thus achieves broader generalization to fit on more diverse new subjects and scenes, similar to image customization over deterministic patch stitching.

In our application, we focus on the one-shot customization case, where only one reference video is provided.
The main challenge in one-shot motion customization is modeling the reference motion without overfitting on the given appearance.
Tuning on multiple videos with the similar motion concept and diverse appearances, the customization module will converge fast on the common information, i.e. the motions, while it learns both spatial and temporal signals with the vanilla diffusion loss when training on a single video.
Leakage of the unwanted appearances into the motion customization module will result in their deterministic reproduction in the generated videos, harming the freedom of synthesizing novel spatial attributes with new prompts.
Leveraging our spatiotemporal disentanglement property of video diffusion models, we develop a targeted training method circumventing these issues to achieve high quality one-shot motion customization with largely simplified tuning modules and pipelines.

\subsection{Timestep Constrained Method}
Prior diffusion-based motion customization methods typically apply LoRA on pre-trained temporal attention layers, and finetune it across all timesteps $t\in[1000, 0]$.
Based on the spatiotemporal disentanglement along timesteps in video diffusion models, where the motion information is primarily processed in early denoising timesteps, we propose to train the temporal LoRA with the groud truth caption in a restricted timestep range $t\in[1000, \tau]$.
$\tau$ is the aforementioned threshold between spatial and temporal signals along the denoising process.
We also constrain the LoRA application during inference within the same timestep range, and at other timesteps the denoising process is proceeded with solely the base model.
The text prompt remains the same new prompt with modified appearances and original motions throughout the inference.

The overall pipeline of our method is illustrated in Fig. \ref{fig:method}.
Compared to previous methods that have to incorporate with auxiliary modules, stages or losses to explicitly debias the appearance learning out of the temporal tuning, our method simplifies the pipeline to only one single temporal LoRA module, one single tuning stage and the vanilla diffusion reconstruction loss.

We also show that our simplified pipeline further facilitates flexible model parameter configurations with stable tuning and consistent performance with minimum appearance leakage.
Furthermore, since our method only constrains the training timesteps, it is very easy to cooperate with other pipelines without any conflict of tuning models or objectives.

\begin{table}[t]
    \centering
    \caption{Ablating different timestep tuning range $\tau$ for one-shot video motion customization, where the base model is tuned at $t\in[1000, \tau]$.
    A smaller $\tau$ corresponds to a wider range of denoising timesteps for finetuning.
    $\tau=1000$ refers to the base model without tuning, and $\tau=0$ refers to tuning the base model at all timesteps.
    The optimal $\tau$ for the downstream task aligns with the peak in our analysis in Fig. \ref{fig:metrics}.
    }
    \begin{tabular}{@{}cc|ccc@{}}
        \toprule
        Base Model & $\tau$ &  
        \begin{tabular}{@{}c@{}}Text \\ Align.$\uparrow$\end{tabular} &
        \begin{tabular}{@{}c@{}}Temp. \\ Const.$\uparrow$\end{tabular} &
        \begin{tabular}{@{}c@{}}Pick \\ Score$\uparrow$\end{tabular} 
        \\
        \midrule
        \multirow{5}{*}{\begin{tabular}{@{}c@{}} ModelScope \\ \cite{modelscope} \end{tabular}}
        & 1000 & 26.05 & 94.88 & 20.13 \\
        & 750 & $\underline{28.04}$ & $\underline{96.39}$ & $20.68$ \\
        & 700 & $\mathbf{28.16}$ & $\mathbf{96.42}$ & $\underline{20.77}$ \\
        & 650 & $27.97$ & $96.31$ & $\mathbf{20.79}$ \\
        & 0 & $27.43$ & $96.25$ & $20.49$ \\
        \midrule
        \multirow{5}{*}{\begin{tabular}{@{}c@{}} Latte \\ \cite{ma2024latte} \end{tabular}}
        & 1000 & $29.28$ & $93.16$ & $20.84$ \\
        & 750 & $31.85$ & $97.12$ & $21.65$ \\
        & 700 & $\mathbf{31.96}$ & $\underline{97.19}$ & $\mathbf{21.68}$ \\
        & 650 & $\underline{31.88}$ & $\mathbf{97.21}$ & $\underline{21.66}$ \\
        & 0 & $31.26$ & $96.99$ & $21.47$ \\
        \midrule
        \multirow{5}{*}{\begin{tabular}{@{}c@{}} CogVideoX \\ \cite{yang2024cogvideox} \end{tabular}}
        & 1000 & $28.15$ & $96.69$ & $20.65$ \\
        & 950 & $\mathbf{30.14}$ & $\mathbf{98.11}$ & $\underline{21.09}$ \\
        & 900 & $\underline{29.93}$ & $\underline{98.10}$ & $21.00$ \\
        & 850 & $29.61$ & $97.76$ & $20.92$ \\
        & 0 & $29.67$ & $97.41$ & $\mathbf{21.30}$ \\
        \bottomrule
    \end{tabular}
    \label{tab:timesteps}
\end{table}

\begin{table}[t]
\centering
    \caption{Comparison with previous SOTA motion customization methods on the TGVE benchmark.
    Our timestep constraining method achieves leading performance without auxiliary modules or stages, and is also compatible to be integrated with existing pipelines.
    $\dagger$ denotes methods that were tested on other datasets and we re-evaluated on the TGVE benchmark for fair comparison.
    $\ddagger$ denotes methods that were tested on other datasets but haven't released code so we cannot re-evaluate.
    }
    \begin{tabular}{@{}lccc@{}}
    \toprule
    Method & \begin{tabular}{@{}c@{}}Text \\ Align.$\uparrow$\end{tabular} & \begin{tabular}{@{}c@{}}Temp. \\ Const.$\uparrow$\end{tabular} & \begin{tabular}{@{}c@{}}Pick \\ Score$\uparrow$\end{tabular} \\
    \midrule
    Tune-A-Video \cite{wu2022tune} & $25.64$ & $92.42$ & $20.09$ \\
    VideoComposer \cite{wang2023videocomposer} & $27.66$ & $92.22$ & $20.26$ \\
    Control-A-Video \cite{chen2023control} & $26.54$ & $92.63$ & $19.75$ \\
    VideoCrafter \cite{video_crafter} & $28.03$ & $92.26$ & $20.12$ \\
    MotionDirector \cite{zhao2023motiondirector} & $27.82$ & $93.00$ & $20.74$ \\
    VMC$^{\dagger}$ \cite{jeong2023vmc} & $25.53$ & $94.58$ & $19.92$ \\
    Gen-1 \cite{esser2023structure} & $28.54$ & $95.77$ & - \\
    MotionClone$^{\dagger}$ \cite{ling2024motionclone} & $27.23$ & $92.88$ & $21.07$ \\
    MotionMatcher$^{\ddagger}$ \cite{wu2025motionmatcher} & $\underline{30.43}$ & $\underline{97.20}$ & - \\
    \hdashline
    Ours-ModelScope & $28.16$ & $96.42$ & $20.77$ \\
    Ours-Latte & $\mathbf{31.96}$ & $97.19$ & $\mathbf{21.68}$ \\
    Ours-CogVideoX & $30.14$ & $\mathbf{98.11}$ & $\underline{21.09}$ \\
    \bottomrule
    \end{tabular}
    \label{tab:comparisons}
\end{table}

\subsection{Experiment Setup}

\paragraph{Base models.}
We implement our training method on three base T2V models: ModelScope \cite{modelscope}, Latte \cite{ma2024latte}, and CogVideoX \cite{yang2024cogvideox}.
All generate videos of 2 seconds and 16 frames, with $256\times256$ resolution for ModelScope, $512\times512$ resolution for Latte, and $480\times480$ for CogVideoX.

\vspace{-.1in}

\paragraph{Datasets.}
To quantitatively evaluate our approach, we apply motion customization on all $76$ videos in the Text-Guided Video Editing (TGVE) competition dataset \cite{loveu} individually.
It is composed of videos from various sources including DAVIS, Youtube and Videovo with various editing tasks such as object, background and style editing.
We use the ground truth captions as the training prompts and sample novel videos for all $4$ editing captions.

\vspace{-.1in}

\paragraph{Metrics.}
We evaluate our generated videos by the following metrics:
Text alignment calculates the CLIP Score \cite{hessel2021clipscore} between the video frames and the new prompts to measure the fidelity of the spatial attributes following the descriptions at inference.
Temporal consistency averages the pairwise CLIP embedding distances between consecutive frames.
Pick Score \cite{kirstain2023pick} trained a model to emulate human preferences of prompt alignment.
Every editing prompt produces $4$ samples, over which the metrics are averaged.

\begin{table}[t]
\centering
    \caption{The top preference rates of our and previous methods in the user study.
    Note that MotionClone is a deterministic approach and thus results in no motion diversity.
    }
    \begin{tabular}{@{}lcc@{}}
    \toprule
    Method & \begin{tabular}{@{}c@{}}Motion \\ Fidelity$(\%)\uparrow$\end{tabular} & \begin{tabular}{@{}c@{}}Motion \\ Diversity$(\%)\uparrow$\end{tabular} \\
    \midrule
    VMC \cite{jeong2023vmc} & $3.8$ & $10.7$ \\
    MotionDirector \cite{zhao2023motiondirector} & $19.4$ & $35.6$ \\
    MotionClone \cite{ling2024motionclone} & $31.8$ & $0$ \\
    \hdashline
    Ours & $\mathbf{45.0}$ & $\mathbf{53.6}$ \\
    \bottomrule
    \end{tabular}
    \label{tab:userstudy}
\end{table}

\subsection{$\tau$ Ablations}

We experiment with choices for the temporal tuning threhshold $\tau$ in our motion customization method.
We present these results in Tab. \ref{tab:timesteps}, using LoRA fine-tuning with a rank and alpha $r=\alpha=4$.
It displays that the optimal $\tau$ consistently align with the peak threshold of the spatiotemporal decomposition property in Fig. \ref{fig:metrics} for each base model.
Meanwhile, the precise value of $\tau$ does not make a significant difference for the final motion customization performance around the optimum, demonstrating the robustness and generalization of our method for practical use.

ModelScope and Latte have separate spatial and temporal attentions in their denoising networks, while ModelScope denoises with U-Net and Latte denoises with transformer.
The overall performance of Latte surpasses ModelScope due to its advanced architecture and larger model size.
CogVideoX is built with unified 3D spatiotemporal attentions, which natively deepen the entanglement of appearance and motion information.
Despite this, our timestep constrained method still achieves leading performance at $\tau=950$ over all other configurations.
This value is significantly larger than other base models as the core motion signals need to be decomposed with a stronger constraint.

In addition, we also list the performance of two baselines for each base model: tuning at all timesteps without a constrained range ($\tau=0$), and the base model without any tuning ($\tau=1000$).
Their performance gaps behind our timestep constrained method indicate the effectiveness of tuning the motion module only at early timesteps, where motion information is dominantly encoded.

\subsection{Comparisons}
We compare our method with various base models at their optimal $\tau$ to other one-shot motion customization approaches that have reported metrics on the TGVE dataset.
The quantitative results are listed in Tab. \ref{tab:comparisons}.
Our motion customization approach yields superior quantitative results to prior SOTAs with a much simplified tuning module and pipeline.
Fig. \ref{fig:comparison} exhibits a visualization of the qualitative comparison.
Our method transfers the reference motion to new subjects and backgrounds with minimal semantic discrepancy compared to other approaches.

\subsection{User Study}

We further conduct an user study to compare motion fidelity and motion diversity of the output videos in the motion customization task, which are ambiguous to measure with automatic metrics.
We compare our method to three previous SOTA approaches under human evaluation: VMC \cite{jeong2023vmc}, MotionDirector \cite{zhao2023motiondirector} and MotionClone \cite{ling2024motionclone}.

In each questionnaire we randomly select $10$ reference videos and their new editing prompts, with two output videos of all 4 methods.
We ask the evaluators to pick the best methods in terms of motion fidelity, which is defined as the temporal similarity between the output and reference videos, and motion diversity, which is defined as the temporal variety between the two output videos.

Our user study involves $30$ participants, each with a random set of questions, and we collected $289$ valid answers in total.
The top pick rates of all methods are listed in Tab. \ref{tab:userstudy}.
Our timestep constrained method outperforms previous SOTAs on both benchmarks.

\subsection{Downstream Extensions}

\paragraph{Ablating Attention Layers.}
Based on our findings of motion disentanglement across timesteps, we are interested in exploring whether motion control can be limited to specific model parameters as well. Given the four query, key, value, and output projections of temporal attention layers, we experiment with restricting training to all possible subsets of these parameters. From our results in Tab. \ref{tab:attention}, we see that only training the value and output projections is necessary for motion customization. In our experiments, we also observe that training only the query and key parameters yields no noticeable change in the generated videos. This suggests that the query and key parameters in temporal attention layers are not responsible for encoding motion information. This allows for cutting the number of trainable parameters in half without sacrificing generation quality.

\begin{table}[t]
    \centering
    \caption{Ablating temporal attention layers with Latte at $\tau=700$.
    By only fine-tuning value and output projections in each attention layer, we cut the number of trainable parameters in half and achieve essentially comparable results.
    }
    \begin{tabular}{@{}c|ccc@{}}
        \toprule
        \begin{tabular}{@{}c@{}} Tunable \\ Layers \end{tabular} &  
        \begin{tabular}{@{}c@{}}Text \\ Alignment$\uparrow$\end{tabular} &
        \begin{tabular}{@{}c@{}}Temporal \\ Consistency$\uparrow$\end{tabular} &
        \begin{tabular}{@{}c@{}}Pick \\ Score$\uparrow$\end{tabular} 
        \\
        \midrule
        Q, K, V, O & $31.69$ & $97.19$ & $21.68$ \\
        V, O & $32.64$ & $97.16$ & $21.62$ \\
        \bottomrule
    \end{tabular}
    \label{tab:attention}
\end{table}

\paragraph{Scaling LoRA Rank and Direct Tuning.}
Prior work usually suffers from increased temporal LoRA rank, as more tunable parameters will more easily overfit on unwanted appearances from the single reference video.
We scale the LoRA rank up to $r=16$.
Moreover, we further extend our method to direct full-parameter fine-tuning.
Previous successful approaches for direct training follow DreamBooth \cite{ruiz2023dreambooth} and require multiple reference samples, as well as a regularization set of general data, to avoid both overfitting on the exemplar appearances or motions.
We instead maintain our settings of only tuning the attention layers on a single reference video, without any additional data.
The direct tuning can be viewed as a full-rank upper bound where the LoRA rank scales to the same as that in the base model.

We present the results in Tab. \ref{tab:rank}.
It contradicts the trivial hypothesis that more parameters always lead to improved one-shot motion customization results.
We attribute this to the limited motion information in a single video, which doesn't need many parameters to model.
On the other hand, this observation also demonstrates the clear spatiotemporal disentanglement of our method, where no appearance is leaked into the tunable module even when much more than necessary parameters are being tuned with the full reconstruction denoising loss, in contrast to traditional DreamBooth pipeline where extra balance data are necessary.

\begin{table}[t]
    \centering
    \caption{Scaling up LoRA ranks and direct full-rank tuning with Latte at $\tau=700$.
    While more tunable parameters contribute marginally to motion customization quality improvement due to limited temporal signals to model in a single video, our spatiotemporal disentanglement property consistently prevent additional parameters from overfitting on the appearance in the reference video.
    }
    \begin{tabular}{@{}c|ccc@{}}
        \toprule
        \begin{tabular}{@{}c@{}}LoRA \\ Rank\end{tabular} &  
        \begin{tabular}{@{}c@{}}CLIP \\ Score$\uparrow$\end{tabular} &
        \begin{tabular}{@{}c@{}}Temporal \\ Consistency$\uparrow$\end{tabular} &
        \begin{tabular}{@{}c@{}}Pick \\ Score$\uparrow$\end{tabular} 
        \\
        \midrule
        $r=\alpha=4$ & $31.69$ & $97.19$ & $21.68$ \\
        $r=\alpha=8$ & $31.61$ & $97.17$ & $21.63$ \\
        $r=\alpha=16$ & $31.34$ & $97.12$ & $21.57$ \\
        All attentions & $31.19$ & $97.23$ & $21.46$ \\
        \bottomrule
    \end{tabular}
    \label{tab:rank}
\end{table}

\section{Conclusion}

We characterize how motion is encoded across timesteps in text-to-video diffusion models by using the trade-off between appearance editing and motion preservation as a timestep-wise probe. This allows us to quantitatively map how motion and appearance compete along the denoising trajectory and to obtain an operational motion-appearance boundary in timestep space that is consistent across diverse architectures. Building on this quantitative analysis, we showed that constraining both training and inference to motion-dominant timesteps simplifies one-shot motion customization framework that achieves high-quality motion transfer without auxiliary modules or tailored losses. These results indicate that timestep-aware, quantitatively grounded scheduling is an effective lever for disentangling and adapting motion in video diffusion models.

\clearpage
{
    \small
    \bibliographystyle{ieeenat_fullname}
    \bibliography{main}
}

\end{document}